%% file: main.tex
\documentclass[11pt]{article}

\usepackage[T1]{fontenc}
\usepackage{lmodern}
\usepackage[margin=1in]{geometry}
\usepackage[round]{natbib}
\usepackage{booktabs}
\usepackage{graphicx}
\usepackage{microtype}
\usepackage{amsmath}
\usepackage{xcolor}
\usepackage[colorlinks=true, linkcolor=blue!60!black, citecolor=blue!60!black,
            urlcolor=blue!60!black]{hyperref}
\usepackage{url}

\graphicspath{{./}}

\newcommand{\pp}{\text{pp}}
\newcommand{\benchrows}{15}

\title{HindsightBench: A Black-Box Behavioral Audit Protocol for\\
Parametric Hindsight in Time-Indexed LLM Decision Tasks}

\author{Haozhe Jia\\
University College Dublin\\
\texttt{haozhe.jia@ucdconnect.ie}}
\date{July 2026}

\begin{document}
\maketitle

\input{sections/00_abstract}
\input{sections/01_intro}
\input{sections/02_related}
\input{sections/03_protocol}
\input{sections/04_panel_artifact}
\input{sections/05_leaderboard}
\input{sections/06_cutoff_colocation}
\input{sections/07_invariances}
\input{sections/08_maintenance}
\input{sections/09_limitations}
\input{sections/10_ethics}

\bibliographystyle{plainnat}
\bibliography{references}

\end{document}

%% file: sections/00_abstract.tex
\begin{abstract}
Large language models leak parametric knowledge of what followed a
historical date into decision tasks indexed by that date --- not
necessarily a lookup of the realized outcome
(\S\ref{sec:patterns}), but knowledge of the period all the same. Existence is settled; what production
users lack is a cheap way to \emph{audit} a given model for it. We present
\textbf{HindsightBench}, a black-box behavioral audit protocol that profiles
parametric hindsight in any time-indexed LLM decision task at probe-level cost
(no backtests, no logprobs, no training-corpus access). The protocol bundles
previously separate ingredients into one preregistered causal design run on a
single task: a four-arm date-manipulation matrix (revealed / date-only /
masked / historically transplanted), dual memory probes (date recovery;
outcome recall), and six per-model metrics --- trigger strength, transplant
effect, post-cutoff placebo, recoverability, behaviorally effective knowledge
cutoff, and a recall--accuracy dissociation coefficient --- with explicit
gates for the metrics whose identifiability is data-dependent. Applying it to \benchrows{} models from seven vendors
on a 258-node vintage-correct macro panel yields three headline patterns:
(i)~the date-trigger reflex is \emph{not a scale phenomenon} --- it tracks
training recency, though what installs it is not identified here: absent
across every 2024 open-weight row where it is measurable, including a 70B
tier with cutoff-aligned recall propensity, present in
every 2026-generation model tested down to 3B active parameters, and
switching on within a single vendor lineage (Qwen3 $\to$ Qwen3.6) inside
one MoE family at a similar ${\sim}$3B-active scale; (ii)~behaviorally effective cutoffs span 22 months
across vendors and precede vendor-reported dates by up to eight months,
which invalidates calendar-window placebo designs and motivates the
model-relative windows the protocol computes; (iii)~audit results are
\emph{not} invariant to serving details --- the trigger estimate loses
mutual-CI stability under BF16 serving of an FP8-referenced model while
AWQ-INT4 preserves it, and one vendor's locked reasoning regime makes an
entire probe non-convergent --- so the protocol ships with operational
requirements (pin quantization, pin the thinking regime, disclose parser and
sampling policy) as part of the benchmark contract. We release the panel,
frozen preregistrations, per-model audit rows with measured dollar costs,
transcripts, and one-command regeneration.
\end{abstract}

%% file: sections/01_intro.tex
\section{Introduction}\label{sec:intro}

An LLM asked to reason about a dated historical decision --- ``it is
September 2008; here is the macro data; generate trading hypotheses'' ---
can silently draw on parametric knowledge of what came next. The existence
of this leak is no longer in question: look-ahead advantages survive entity
anonymization \citep{glasserman2023}, are exposed by events unpredictable
ex ante \citep{sarkar2024}, inflate in-sample forecasts \citep{liang2026}, and decay
LLM trading-agent Sharpe ratios past the training cutoff by half
\citep{profitmirage2025}. What a practitioner deploying a specific model on a
specific time-indexed task still cannot do is cheaply answer four operational
questions: \emph{Does my model exhibit hindsight on this task? What triggers
it? Where does the model's behavioral knowledge actually end? And do those
answers survive my serving configuration?}

HindsightBench is a protocol for answering these questions by black-box
behavioral measurement alone. It requires only API access at probe-level
cost --- a full audit row for a mid-tier commercial model cost us
\$19--30 at 2026 list prices (\S\ref{sec:leaderboard}) --- and no backtest
infrastructure, no logprobs, and no knowledge of the training corpus.

\paragraph{What is new.} Every individual ingredient of the protocol has a
predecessor, and we position against each occupant explicitly
(\S\ref{sec:related}). Paired true/fake-date prompts exist
\citep{fakedate2026}; date-only recall probes exist \citep{gao2025lap};
masking designs exist \citep{glasserman2023,knowingdoing2026}; effective-cutoff
probing exists at the resource level \citep{dateddata2024} and the freshness
level \citep{llmlagbench2025,cutoffbench2025}. Our claims are three:

\begin{enumerate}
\item \textbf{An integrated causal protocol.} The components are chained
into a single attribution design on one task: the four-arm matrix separates
the \emph{trigger} (the date token) from date-identifying context and from
data content; the transplantation arm tests whether a remembered narrative
\emph{relocates} to an asserted date; the dual probes separate what is
\emph{recoverable} from what is \emph{recalled} and both from what is
\emph{behaviorally active}; and identifiability gates state when the
data-dependent metrics (placebo, recoverability, cutoff, dissociation) are
defined. No prior work runs this chain end-to-end, which is what
allows per-model attribution rather than per-model detection.
\item \textbf{A multi-metric hindsight profile leaderboard.} \benchrows{}
models from seven vendors, each summarized in one audit row of six metrics
under a frozen protocol with per-model deviations disclosed. Existing
model sweeps are single-axis: freshness \citep{llmlagbench2025}, numeric
recall \citep{numleak2026}, cutoff-instruction compliance
\citep{simulatedignorance2026}. The profile view is what surfaces our
headline pattern --- the trigger reflex is not a scale phenomenon but
tracks training recency, and switches on \emph{within} a vendor lineage (Qwen3 $\to$ Qwen3.6) inside one
MoE family at a similar ${\sim}$3B-active scale.
\item \textbf{In-task cutoff co-location.} The behaviorally effective
knowledge cutoff is measured \emph{on the same task and stimuli} as the
hindsight profile, rather than on a separate freshness instrument. This is
the surviving role of the cutoff locator --- two deployed tools already
locate cutoffs in general \citep{llmlagbench2025,cutoffbench2025} --- and
it matters because placebo windows, post-cutoff safety claims, and
train/test splits are only valid relative to the boundary of the task at
hand (\S\ref{sec:cutoff}). Measured cutoffs span 22 months across our
vendors and precede vendor-reported dates by up to eight months.
\end{enumerate}

\paragraph{A fourth, unglamorous contribution} is a set of measured
\emph{protocol invariance results and operational requirements}
(\S\ref{sec:invariances}): the audit's headline metric is \emph{not} stable
under serving quantization changes (BF16 vs.\ the FP8 reference on the same
weights breaks mutual-CI containment; AWQ-INT4 does not), and one vendor's
provider-locked reasoning regime makes an entire probe non-convergent.
A benchmark of behavioral reflexes must therefore pin serving details as
part of its contract, and we specify which ones, with evidence.

We release the vintage-correct panel, all frozen preregistrations (with
SHA-256 hashes), per-model transcripts and audit rows, measured per-row
dollar costs with an explicit gap ledger, and one-command regeneration of
all leaderboard-derived tables and figures from frozen row
files.\footnote{\url{https://github.com/Khaozhe/hindsightbench}}

%% file: sections/02_related.tex
\section{Related Work}\label{sec:related}

We cut the literature along three lines, then list the component occupants
our protocol integrates.

\paragraph{End-to-end decay.} One line measures hindsight by its downstream
footprint: portfolio performance that collapses past the training cutoff.
Look-Ahead-Bench standardizes the measurement of look-ahead bias in
point-in-time financial LLMs via alpha decay across temporally distinct
market regimes \citep{lookaheadbench2026}; Profit Mirage audits five LLM
trading-agent frameworks and finds 51--62\% Sharpe decay past cutoff
\citep{profitmirage2025}; \citet{liang2026} quantifies the pre/post-cutoff
forecast-error discontinuity; \citet{didisheim2025} tie memorization strength
to data frequency, aggregation, and scale. These designs need a full
trading-evaluation stack, and they observe model and evaluation harness
jointly --- they detect that leakage happened, not what triggered it.
Look-Ahead-Bench's own positioning explicitly sets itself apart from
approaches that ``primarily test inner lookahead knowledge via Q\&A''; that
is the space this protocol occupies.

\paragraph{Detection with white-box access.} A second line detects
memorization statistically: membership-inference-style probes and
recall-interaction regressions requiring logprobs \citep{gao2025lap}. Our
adaptation of the
recall probe (\S\ref{sec:protocol}) is black-box --- a frequency
approximation over repeated sampling --- and our benchmark shows its
identifying assumption (contamination as accuracy gain) is model-conditional,
which detection pipelines calibrated on one model inherit silently.

\paragraph{Freshness-level cutoff location.} \citet{dateddata2024}
introduced the effective-vs-reported cutoff distinction at the resource
level; LLMLagBench locates training-data boundaries behaviorally via
news-event QA with changepoint detection on 35 models
\citep{llmlagbench2025}; CutoffBench deploys an
independent estimator as a web tool \citep{cutoffbench2025}. We do not
compete on general freshness: our locator is \emph{in-task} --- same
stimuli, same task as the audit --- because the boundary that matters for a
placebo window or a train/test split is task-relative
(\S\ref{sec:cutoff}).

\paragraph{Component occupants.} Paired true/fake-date macro prompts:
\citet{fakedate2026} (distributional test against far-future dates; our
transplant arm asserts \emph{historical} dates and reads a directional
narrative field). Anonymization/masking: \citet{glasserman2023}, and a
memory-controlled benchmark for LLM trading agents on Chinese equities
(CSI300): \citet{knowingdoing2026}. Cutoff-instruction compliance: simulated
ignorance fails systematically --- cutoff instructions leave a 52\%
performance gap against true ignorance \citep{simulatedignorance2026},
consistent with \citet{lopezlira2025}'s finding that instructed forgetting
fails --- our complement is that date \emph{assertion} is obeyed almost
without question.
Ex-ante inference under temporal constraints: \citet{exante2025}. Numeric
benchmark leakage profiling: \citet{numleak2026}. Vintage blur in macro
recall: \citet{fedtotalrecall2025} --- addressed here by a vintage-correct
panel. Mitigation --- chronologically consistent pretraining
\citep{chrono2025,ittchrono2025,datedgpt2026}, counterfactual-anchored
decoding \citep{fincad2026}, and inference-time logit adjustment against a
pair of forget/retain models \citep{logitadjust2025} --- solves a problem
whose per-model diagnosis is our subject; an audit row tells a deployer
whether they need these tools at all.

\paragraph{Audit-methodology lineage.} As an audit instrument, the
protocol sits at the model layer of the governance/model/application
taxonomy of \citet{mokander2023auditing}. \citet{casper2024blackbox}
argue that black-box access alone is insufficient for rigorous AI audits;
our serving-precision results (\S\ref{sec:quant}) supply empirical
support from the opposite direction --- even the black-box measurement
itself is unstable unless deployment details are pinned --- and the
protocol's identifiability gates are the instrument-level response:
metrics that stop being identified are reported as undefined rather than
estimated.

%% file: sections/03_protocol.tex
\section{The BM-1 Protocol}\label{sec:protocol}

\begin{figure*}[tbp]
\centering
\includegraphics[width=0.98\textwidth]{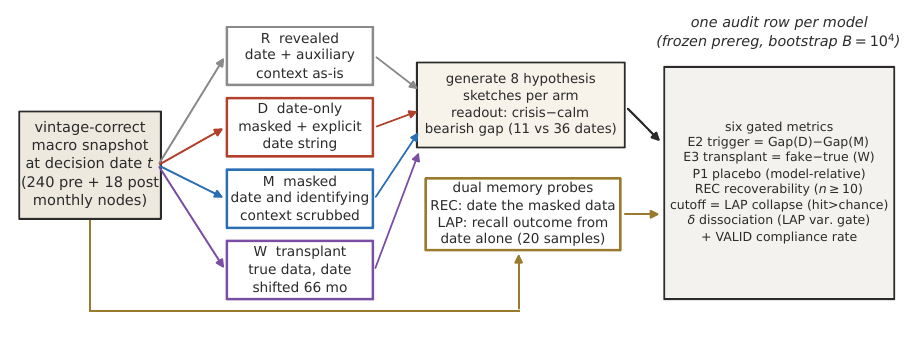}
\caption{Protocol overview. Each vintage-correct decision node fans out
into four arms that manipulate only the dating information; the same nodes
feed two memory probes. Arms and probes jointly identify the six metrics
and their identifiability gates, producing one audit row per model under a
hash-frozen preregistration.}
\label{fig:protocol}
\end{figure*}

One audit row per model (Figure~\ref{fig:protocol}); \emph{BM-1} is the
protocol's preregistration codename (experiment codenames --- BM-$\ast$
here, GD-$\ast$ for the cross-domain checks --- index the hash-frozen
freeze registry shipped with the artifact). The arm/probe collection protocol and primary estimands were frozen (SHA-256
\texttt{fbcdffc1}) before any non-development model was called;
identifiability gates and parser corrections introduced after instrument
diagnostics are versioned, disclosed, and applied uniformly to all eligible
rows (\S\ref{sec:maintenance}). An English
companion of the preregistration ships with the artifact. Arm-contrast
estimates use paired bootstrap CIs ($B=10{,}000$, fixed seed); the
dissociation coefficient uses a HAC regression $t$-statistic
(\S\ref{sec:metrics}); probe rates are raw shares. Cross-model inference
is deliberately restricted to descriptive ranking and group contrasts ---
no pairwise significance forest.

\subsection{Task and arms}

The carrier task is macro hypothesis generation on a vintage-correct panel
(\S\ref{sec:panel}): given a monthly macro snapshot as of decision date $t$,
produce eight structured hypothesis sketches with direction calls. Four
arms manipulate only the dating information:

\begin{itemize}
\item \textbf{R (revealed)}: date and auxiliary context as-is;
\item \textbf{D (date-only)}: the masked prompt plus the explicit date
string --- isolating the date token as a channel;
\item \textbf{M (masked)}: date removed and date-identifying auxiliary
context scrubbed;
\item \textbf{W (wrong date / transplant)}: true data relabeled with a
date shifted 66 months (cyclically); a companion 72-month
\emph{month-preserving} variant controls the seasonal confound: its
fake-keyed gap is $+18.4\pp$, $6.9\pp$ below the 66-month arm's $+25.3\pp$
and inside the preregistered $10\pp$ shift-family band, which is defined on
that gap; under the E3 definition below the corresponding fake-minus-true
contrast is $+16.6\pp$ [5.4, 27.7]. If a crisis
narrative follows the \emph{asserted} date onto mismatched data, memory of
the period, not properties of the data, drives the behavior.
\end{itemize}

The behavioral readout is the \emph{crisis--calm bearish gap}: the
difference in bearish-call share between a preregistered crisis set (11
dates, an ex-ante event list --- six global-financial-crisis months
2008-09 through 2009-02, COVID 2020-03/04, and the 2022 inflation bear
2022-06/09/10 --- not the output of any recession indicator) and a
calm-year set (all 36 dates of 2013/2014/2017). Membership enters at
analysis time only; prompts are partition-independent, and a post-freeze
perturbation ablation (leave-one-event-out, drop-one-date jackknife,
alternative calm years, balanced counts) flips no frozen-significant
row's sign --- 6/208 flipped cells, all on rows statistically
indistinguishable from zero, and the CI keeps excluding zero in 57/60
leave-one-event-out cells. Scoring reads only the structured direction
field: W-arm free text does retrieve the asserted period ---
fake-date-year mentions reach 86--100\% of completions on the most
affected rows, recognizable event references up to 77\% --- which is the
transplant mechanism itself made visible; a 15-model transcript audit
finds true-date-year references rare ($\leq 6\%$ per row,
fake-vantage-consistent on inspection) and dated strings essentially
absent from masked-arm text (one occurrence in 6{,}396 full-tier masked
completions). Perturbation and transcript-audit outputs ship with the
artifact.

\subsection{Probes}

\textbf{Date recovery (REC)}: shown the masked snapshot, the model is asked
to date it (temperature 0); scored per row as exact year-month accuracy,
calendar-year accuracy, and median absolute month offset, pre- and
post-cutoff; the raw probe output also supports coarser readings, such as
the within-$\pm$12-months rate the companion analysis headlines for its
primary model. Tables print calendar-year accuracy. \textbf{Outcome
recall (LAP)}: a date-only query for the realized outcome, adapted from
\citet{gao2025lap} to the black-box setting as a frequency approximation
over 20 samples at temperature 1.0. LAP per month yields both a recall rate
and --- through its collapse --- the effective cutoff. Aggregating the
sampled directions per month also yields the \emph{recall hit rate}: the
share of pre-cutoff months whose net recalled direction matches the
realized outcome --- the guessing gate for the cutoff metric below.

\subsection{Six metrics and their gates}\label{sec:metrics}

\begin{enumerate}
\item \textbf{E2 (date trigger)} $= \mathrm{Gap}(D) - \mathrm{Gap}(M)$.
\item \textbf{E3 (transplant)} $= \mathrm{Gap}_{\text{fake}}(W) -
\mathrm{Gap}_{\text{true}}(W)$: the same W generations scored once under
the asserted (fake) date labels and once under the true ones.
\item \textbf{P1 (post-cutoff placebo)} $= D - M$ bearish-share difference
on post-cutoff dates, where no memory exists to trigger. \emph{Gate}: if
the measured cutoff falls inside the calendar placebo window, the window is
recomputed model-relatively (\S\ref{sec:cutoff}); both computations are
frozen per model.
\item \textbf{REC (recoverability)}. \emph{Gate}: a window is reported only
with $n \geq 10$ convergent probe answers (one vendor's reasoning mode
fails this row-wide, \S\ref{sec:invariances}; one reduced-tier row fails it
in both windows).
\item \textbf{Effective cutoff}: the last month with LAP $> 0.1$ in the
monthly LAP series. \emph{Gate}: defined only when the recall hit rate
exceeds chance --- for three small models the probe measures guessing, and
their never-collapsing ``cutoffs'' are artifacts, reported as undefined.
\item \textbf{$\delta$ (dissociation)}: the signal $\times$ LAP interaction
in a correctness regression (HAC, lag 6) --- a \emph{signal-conditional
recall--correctness interaction}: its sign records whether the
LAP--correctness association differs between bullish and bearish signals,
not by itself whether recall globally loads on accuracy rather than
narrative. \emph{Gate}: undefined when LAP variance on the
regression's estimation dates is $\leq 10^{-4}$ (interaction collinear
with signal; the degenerate regression once produced $\delta = 12.5$
before gating). Near-saturated rows --- LAP $\approx 1$ on every
estimation date --- fall under this gate; three cells in
Table~\ref{tab:rows} are gated this way, one of them as a disclosed
post-freeze correction.
\end{enumerate}

\textbf{VALID} (schema-compliant generation share) is reported for every
row; models that cannot carry the task exit with the record kept rather
than being silently dropped.

\subsection{Does the protocol transfer across tasks?}\label{sec:transfer}

The protocol is a transferable template for time-indexed directional
tasks---each domain must re-specify the direction semantics and the
behavioral contrast. As evidence, a reduced
65-date audit on a second domain (10-year Treasury yield direction, sign
pinned so that bearish $=$ yields fall) reproduced the trigger and
transplant effects cross-model with asset-appropriate signs: E2/E3 both
significant on two of three models tested (up to $+63.2\pp$
[45.9, 80.0] trigger and $+64.3\pp$ [23.6, 104.2] transplant), the third
positive but with a structurally wide transplant CI under the reduced
window (\S\ref{sec:crossdomain}). Domain-transfer requires re-pinning the
direction semantics and re-deriving the crisis/calm sets; nothing else
changes.

%% file: sections/04_panel_artifact.tex
\section{Panel and Artifact}\label{sec:panel}

\paragraph{Panel.} 258 monthly decision nodes: 240 pre-cutoff
(2005-01--2024-12) and 18 post-cutoff (2025-01--2026-06, refreshed on a
rolling basis; \S\ref{sec:maintenance}). Of the 18, the calendar placebo
window (P1, \S\ref{sec:metrics}) evaluates the 17 dates from 2025-02:
2025-01 is the reference model's vendor-reported cutoff month and is
excluded as a boundary month (preregistered), while the LAP locator scans
all 18. Each node is a vintage-correct
macro snapshot assembled from archival data vintages (ALFRED), so the
information set at node $t$ contains exactly what was published by $t$ ---
this removes the cross-vintage blur that \citet{fedtotalrecall2025} show
LLMs themselves exhibit, and makes the masked arm meaningful: what remains
after masking is a historically faithful, date-scrubbed information set.

\paragraph{Licensing.} The underlying vintage series are public-domain
(U.S.\ federal statistical releases via ALFRED). Market-derived direction
labels are released in conservative aggregate form (direction and coarse
magnitude bins, not raw index levels).

\paragraph{What ships.} (i) The panel snapshots and arm-transformation
code; (ii) every frozen preregistration with SHA-256 hashes and freeze
timestamps, including the data-dependent extensions with their decision
rules; (iii) full per-model transcripts for all arms and probes;
(iv) per-model audit-row JSONs --- the single numeric source from which the
leaderboard table, the CSV export, and both papers' table files are
generated by one script with a drift-check mode; (v) the usage ledgers and
the cost-attribution script behind the measured-cost column; (vi) runner
and analysis code with per-model adapter notes. The artifact is the
versioned public repository at
\url{https://github.com/Khaozhe/hindsightbench}, with the v1.0 release
archived on Zenodo (DOI
\href{https://doi.org/10.5281/zenodo.21453191}{10.5281/zenodo.21453191});
Croissant (core + RAI) metadata and dataset hosting are planned for the
archival submission.

\paragraph{Reproducibility discipline.} Every leaderboard number and every
leaderboard-derived table or figure is generated from frozen row files by
scripts in the artifact; all other reported quantities are script-generated
from frozen experiment outputs or usage ledgers. The leaderboard
generator has a \texttt{--check} mode that fails if any published table
drifts from the frozen rows, and the same generated row block is included
by both this paper and the companion findings paper, so the two cannot
silently diverge.\footnote{The companion findings paper --- the causal
anatomy on the equity task from which this protocol was abstracted --- is
not yet publicly available. Every quantity this paper attributes to
it is regenerated from the frozen artifact released here, so no claim in
this paper depends on reading it; where its argument is load-bearing
(\S\ref{sec:patterns}) we reproduce that argument inline rather than
defer to it.}

%% file: sections/05_leaderboard.tex
\section{The \benchrows-Model Leaderboard}\label{sec:leaderboard}

Table~\ref{tab:rows} gives the audit rows for \benchrows{} models from seven
vendors (OpenAI, Anthropic, Google, DeepSeek, Alibaba/Qwen, Moonshot, Meta),
spanning three training generations and 1B--70B(+) parameters. The row block
is machine-generated from the frozen per-model row files (identical, by
construction, to the companion paper's appendix table).

\begin{table*}[t]
\centering
\scriptsize
\setlength{\tabcolsep}{3.5pt}
\resizebox{\textwidth}{!}{%
\begin{tabular}{@{}llllllllll@{}}
\toprule
Model & Gen & E2 & E3 & P1 & REC pre/post & LAP pre/post (hit) & Cutoff & $\delta$ (t) & VALID \\
\midrule
GPT-5.5 & 2026 & +.215 [.10,.33] & +.161 [.09,.23] & $-$.073\textsuperscript{m} & 100\%/94\% & .999/.161 (100\%) & 2025-06 & ---\textsuperscript{v} & 100\% \\
Claude Sonnet 5 & 2026 & +.326 [.19,.46] & +.450 [.26,.64] & $-$.062\textsuperscript{m} & 95\%/69\% & 1.000/.556 (92\%) & 2025-10 & ---\textsuperscript{v} & 100\%\textsuperscript{p} \\
Kimi-K2.6 & 2026 & +.332 [.25,.41] & +.385 [.26,.51] & $-$.062\textsuperscript{m} & ---\textsuperscript{r} & .330/.008 (98\%) & 2025-02 & +.254 (2.99) & 100\% \\
Qwen3.6-35B-A3B & 2026 & +.237 [.12,.35] & +.346 [.24,.46] & $-$.029 & 26\%/0\% & .982/.000 (78\%) & 2024-12 & +.514 (1.19) & 100\% \\
Qwen3.6-27B & 2026 & +.193 [.11,.28] & +.181 [.11,.25] & $-$.055 & 9\%/0\% & .919/.000 (72\%) & 2024-11 & +.271 (1.23) & 100\% \\
GPT-5.4-mini & 2026 & +.144 [.07,.22] & +.187 [.08,.29] & +.014\textsuperscript{m} & 49\%/18\% & 1.000/.414 (74\%) & 2025-09 & ---\textsuperscript{v} & 100\%\textsuperscript{p} \\
Claude Haiku 4.5 & 2025 & +.383 [.30,.47] & \textbf{+.495} [.34,.64] & $-$.007 & 20\%/0\% & .838/.000 (69\%) & 2024-10 & +.162 (1.35) & 100\% \\
DeepSeek v4-flash & 2025 & +.416 [.30,.54] & +.309 [.14,.48] & +.040\textsuperscript{m} & 95\%/65\% & .815/.103 (89\%) & 2025-07 & +.019 (0.13) & 100\% \\
Gemini 2.5 Flash & 2025 & +.256 [.14,.37] & +.311 [.20,.43] & $-$.025 & 80\%/6\% & .337/.000 (94\%) & 2024-06 & \textbf{$-$.266} ($-$2.03) & 100\% \\
Gemini 2.5 Pro & 2025 & +.167 [.10,.24] & +.242 [.13,.36] & ---\textsuperscript{c} & 96\%/33\% & .922/.000 (98\%) & 2024-05 & +.235 (0.29) & 100\% \\
Qwen3-30B-A3B\textsuperscript{n} & 2025 & +.063 [$-$.10,.24] & +.016 [$-$.14,.18] & +.066 & 18\%/0\% & .807/.000 (67\%) & 2024-04 & $-$.133 ($-$1.11) & 100\% \\
Llama 3.1 70B & 2024 & $-$.047 [$-$.12,.03] & +.008 [$-$.10,.11] & +.004 & 18\%/0\% & .854/.000 (\textbf{68\%}) & 2023-12 & +.007 (0.13) & 100\% \\
Llama 3.1 8B & 2024 & +.056 [$-$.10,.20] & +.076 [$-$.02,.18] & +.051 & 5\%/0\% & .954/.144 (50\%)\textsuperscript{g} & ---\textsuperscript{g} & ---\textsuperscript{g} & 100\% \\
Llama 3.2 3B & 2024 & ---\textsuperscript{k} & ---\textsuperscript{k} & ---\textsuperscript{k} & 5\%/0\% & .921/.544 (48\%)\textsuperscript{g} & ---\textsuperscript{g} & ---\textsuperscript{g} & 52\% \\
Llama 3.2 1B & 2024 & $-$.096 [$-$.24,.05] & ---\textsuperscript{k} & +.055 & ---/--- & .522/.450 (43\%)\textsuperscript{g} & ---\textsuperscript{g} & ---\textsuperscript{g} & 79\% \\
\bottomrule
\end{tabular}}
\caption{Audit rows in benchmark-table order (by training generation).
REC prints calendar-year accuracy; an em-dash in REC marks a window below
the $n \geq 10$ convergence gate (\S\ref{sec:metrics}) --- Kimi-K2.6
row-wide\textsuperscript{r}, and Llama 3.2 1B in both windows (0 convergent
pre-cutoff answers, 1 post-cutoff). Boldface marks the three cells read as
headline evidence in \S\ref{sec:patterns}: the largest transplant effect,
the significantly negative $\delta$, and the recall hit rate of the
no-trigger model whose recall collapses at its reported cutoff.
\textsuperscript{m}~model-relative placebo window (measured cutoff falls
inside the calendar window; both computations frozen per model).
\textsuperscript{n}~row added after the fourteen-row freeze from a
preregistered pilot's GO/KILL stage; BM-1-compatible protocol; the
date-recovery probe was deferred at row freeze and backfilled under the
identical serving configuration (no other cell changed); inclusion is
data-dependent and disclosed.
\textsuperscript{c}~post-cutoff arms not generated (cost decision).
\textsuperscript{r}~probe non-convergent under the provider-locked reasoning
regime (\S\ref{sec:invariances}). \textsuperscript{v}~$\delta$ undefined:
LAP variance $\leq 10^{-4}$ on the regression's estimation dates
(interaction collinear with signal). The GPT-5.5 cell printed $+.065$
($t=2.96$) in its initially frozen row, which predated the variance gate;
on its 152 estimation dates LAP is exactly saturated (variance zero ---
the one sub-saturation month is excluded by the regression's own sample
rule), so the gated pipeline voids the cell; disclosed as a post-freeze
correction rather than silently regenerated.
\textsuperscript{p}~parser v2 recovery (uniform trailing-comma
normalization; disclosed in \S\ref{sec:invariances}).
\textsuperscript{g}~recall hit rate at chance: cutoff and $\delta$
undefined.
\textsuperscript{k}~unidentified under the sample-parity gate: under partial
schema compliance the two sides of a contrast return usable generations on
\emph{different} dates, so the contrast must be restricted to the dates they
share --- and that surviving intersection is selected by attrition rather
than by design, on the outcome. A contrast is reported only when both sides
cover the full preregistered windows. On Llama 3.2 3B (52\% compliance) this
gates E2 (shared window 3 of 11 crisis months; D is bearish on .542 of the
shared crisis months against .333 of those only D kept, while M is .125
against .375, both calm rates flat), E3 (the true-date and fake-date keyings
of the W arm share \emph{no} crisis month), and the placebo (D covers 8
post-cutoff dates, M 11, sharing 6). On Llama 3.2 1B (79\%) it gates E3
(4 of 11 crisis months shared). All are disclosed post-freeze corrections
(\S\ref{sec:maintenance}): the superseded cells read $+.252$ $[-.05,.51]$,
$-.520$ $[-.70,-.30]$, $-.033$ and $-.162$ $[-.35,-.01]$; date-matching
gives $+.486$ $[+.25,+.75]$, no value, $-.125$ and $-.160$ $[-.35,+.01]$.
The instability across that choice is why the cells are gated rather than
reprinted --- and the E2 correction moves \emph{against} the null reading
of this cohort. Protocol tiers and quantization per row are
recorded in the artifact registry.}
\label{tab:rows}
\end{table*}

\begin{figure*}[tbp]
\centering
\includegraphics[width=0.92\textwidth]{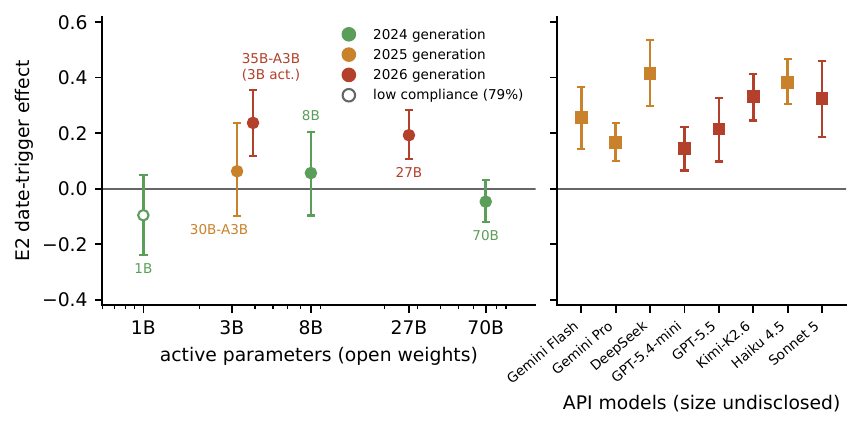}
\caption{E2 date-trigger effect (95\% bootstrap CI) against parameter
scale, colored by training generation. Left: open-weight models at known
active-parameter counts. Right: API models (sizes undisclosed). No
2024-generation estimate is distinguishable from zero at any scale; Llama
3.2 3B is absent from the panel because its E2 is
unidentified\textsuperscript{k} rather than null, so plotting a point for
it would overstate what the row supports; every
2026-generation model triggers; within the same vendor's MoE family at a similar ${\sim}$3B-active scale the
2025-generation Qwen3 is null while the 2026-generation Qwen3.6 is
significantly positive.}
\label{fig:scalegen}
\end{figure*}

\subsection{Core patterns}\label{sec:patterns}

\paragraph{Not a scale phenomenon.} Significant
date-trigger (E2) and transplant (E3) effects appear in 6/6 tested
2026-generation models --- including a 3B-active MoE and the cheapest tier
of its family --- in 4/5 of the 2025 cohort, and in 0/3 of the 2024
open-weight rows where the trigger is measurable at all
(Figure~\ref{fig:scalegen}). The fourth 2024 row, Llama 3.2 3B, is not
counted as a null: at 52\% schema compliance its two arms do not cover the
same dates, and E2 is gated as unidentified\textsuperscript{k}. A second
instrument says the same thing without reference to any threshold: the
recall probe puts this row's outcome-recall hit rate at \textbf{48\%} ---
chance --- alongside 50\% on the 8B and 43\% on the 1B, against 68\% on the
70B. Because the LAP gate measures outcome-direction recall rather than
period-semantic memory, chance-level hit rates make these rows uninformative
for the outcome-memory interpretation of E2; they do not rule out
date-triggered period semantics. The
honest reading is that the smallest tiers carry no evidence either way, and
the negative claim rests on Llama 3.1 70B, whose outcome-recall
propensity collapses exactly at Meta's reported cutoff, with no trigger. Under the parity gate \emph{no} 2024 interval excludes zero in
either direction --- the two negative E3 cells that previously did were
both sample-mismatch artifacts of the reduced tier. We state the other
direction plainly, because the gate is what suppresses it: date-matching
the 3B row's E2 gives $+.486$ $[+.25,+.75]$, which would be the
\emph{largest} E2 in the panel and the only 2024 cell pointing at a
trigger. We do not report it, not because it is inconvenient but because
its window is chosen by which dates survived: on the shared crisis months
D is bearish .542 against .333 on the months only D kept, and M is .125
against .375, while both arms' calm rates are flat. A reader who wants the
cohort's negative result to rest on something should read Llama 3.1 70B,
not the absence of this cell. The
decisive observation is \emph{within-vendor}: Qwen3-30B-A3B (2025) has
cutoff-aligned recall (measured cutoff 2024-04) but no trigger
(E2 $+6.3\pp$ $[-9.7, +23.6]$; E3 $+1.6\pp$ $[-14.1, +17.5]$), while
Qwen3.6 MoE --- same vendor, same MoE family, a similar ${\sim}$3B-active budget ---
triggers at $+23.7\pp$. Whatever installs the reflex changed between those
two training pipelines within a narrow active-parameter band. One attribution confound is
explicit, and it is worse than collinearity. Ordered by measured cutoff,
trigger status steps exactly once across the \emph{twelve recall-capable}
rows (absent through 2024-04, present from 2024-05 onward; the three
recall-gated rows have no measurable cutoff and are excluded from the
ordering). On those twelve, ``measured cutoff $\geq$ 2024-05'' separates
trigger status with zero errors while ``2025-or-later generation'' makes
one --- Qwen3-30B-A3B, a 2025 model on the non-trigger side, which is also
the row our within-vendor contrast leans on. \emph{Corpus period is
therefore the better in-sample predictor of the two}, and we do not claim
otherwise. Three considerations keep the pure corpus-period reading from
being decisive, without settling attribution: boundary adjacency (one
measured month separates the last non-trigger from the first trigger, so
that account must load the whole reflex onto one month of additional
text); the corpus fork (the measured collapse is behavioral, not a corpus
clock --- if Qwen3-30B's corpus ends near its measured 2024-04 the
adjacency argument applies in full, and if it runs later, then a model
trained past the boundary failed to acquire the reflex, which contradicts
the account directly); and memory-without-reflex (Llama 3.1 70B's recall
propensity collapses at its reported cutoff, with no trigger at all).

\paragraph{Memory without trigger; trigger without reconstruction.} Two
models (Llama 3.1 70B; Qwen3-30B-A3B) carry cutoff-aligned recall
propensity that never activates behaviorally --- and both sit at
date-recovery rates that round to the same 18\%: weak reconstruction, cutoff-aligned
recall, no reflex. A post-freeze per-date cross-tabulation sharpens the
dissociation: dates of high generic recall show no trigger elevation in
any of four models examined; elevation concentrates where recall
coincides with crisis semantics. Conversely, the trigger
does not require date-reconstruction ability: Qwen3.6-27B recovers the date
of masked data in only 9\% of months yet triggers at $+19.3\pp$, and the
largest transplant effect in the table belongs to a model with 20\%
recovery (Claude Haiku 4.5, E3 $+49.5\pp$; its sibling Sonnet 5 is second
at $+45.0\pp$). Elicitable, recalled, and behaviorally
active are three different properties; single-axis audits conflate them.

\paragraph{The dissociation coefficient has both signs.} $\delta$ is
significantly negative on Gemini 2.5 Flash ($-0.266$, $t=-2.03$) and
significantly positive on Kimi-K2.6 ($+0.254$, $t=2.99$, well-identified)
--- a signal-direction-specific heterogeneity across models. The
interaction's sign does not by itself decide whether an accuracy-calibrated
probe detects contamination; pipelines should treat $\delta$ as a per-model
measurement, not an assumption.

\paragraph{Which channel dominates is also per-model.} The four-arm matrix
separates the date token from the date-identifying context, but the six
gated metrics report only the date-token contrast; the context contrast
$E1 = \mathrm{Gap(R)} - \mathrm{Gap(D)}$ is computable on every row from
arms we already ship, and we had not reported it. Estimated post hoc under
the same estimator and windows
(\texttt{analyze\_bench\_e1.py}), E1 is null on \textbf{13 of 15} rows ---
so ``the auxiliary context adds nothing once the date is present'' holds
broadly --- with two significant exceptions of opposite sign: Kimi-K2.6
$-11.2\pp$ [$-18.8$, $-4.2$] and \textbf{GPT-5.4-mini $+19.4\pp$}
[$+11.3$, $+27.1$], the latter \emph{exceeding that model's own E2}
($+14.4\pp$). On one of fifteen models the dominant leakage channel is the
context, not the date. This is the four-arm matrix earning its extra arm.
A two-arm revealed-versus-masked design --- the standard construction in
existing masking work --- would have recovered GPT-5.4-mini's total
leakage (33.8pp) and assigned all of it to the date token; the split
attributes 57\% of it to the context instead. An audit can therefore get
the magnitude roughly right and the mechanism entirely wrong, which is a
failure no amount of precision on the total will surface. The
discrimination is also free: the revealed arm is already generated for
every row, so E1 costs no additional inference. It does carry one
operational requirement, which we learned by getting it wrong: E1 must be
computed \emph{within batch}, on the replicates the two arms share. One of
our rows ships R at one replicate and D at three, and pooling them shifts
that row by 6.5pp --- replicate noise entering as an apparent channel
effect. We report E1 here as a post-freeze addition rather than a seventh
gated metric, and promote it to a gated metric, with the parity rule, in
the next protocol version.

\subsection{Measured cost}\label{sec:cost}

Costs are ledgered per call and attributed mechanically (fixed time windows
with exact expected call counts, asserted). Representative full-protocol
rows with repo-frozen rates: Claude Haiku 4.5 \$29.45
(10.5M tokens), GPT-5.4-mini \$19.30 (8.5M); GPT-5.5's premium output
pricing makes it the most expensive row at $\geq$\$115 (output-side lower
bound). The ledger states its gaps explicitly: one API row predates the
ledger entirely; several rows carry exact token counts but no repo-frozen
billing rate (reported tokens-only); the batch-run row's tokens are
harvested from the retrievable batch results but await a billing-anchored
rate; self-hosted rows are GPU-rental time. An audit row is, in every case,
orders of magnitude cheaper than a backtest-grade evaluation stack.

%% file: sections/06_cutoff_colocation.tex
\section{In-Task Cutoff Co-Location}\label{sec:cutoff}

\begin{figure*}[tbp]
\centering
\includegraphics[width=0.92\textwidth]{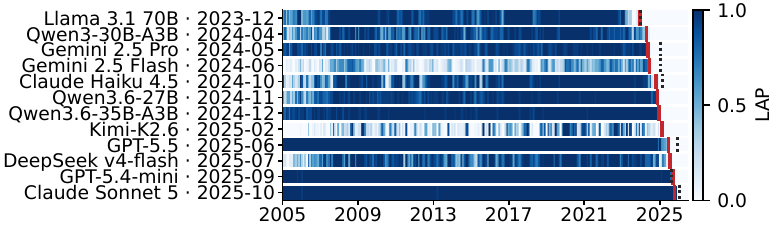}
\caption{Monthly LAP (outcome-recall propensity) per model, rows ordered by
measured cutoff; the collapse from color to white \emph{is} the behaviorally
effective cutoff (red tick), against the vendor-reported date (dotted black
tick) where one is published. Guessing-tier models are excluded (cutoff
undefined). The right edge forms a 22-month staircase.}
\label{fig:staircase}
\end{figure*}

The effective cutoff --- the last month at which the model's date-only
outcome recall stays above threshold \emph{on this task's stimuli} --- is a
by-product of the LAP probe, not a separate instrument. Three findings
(Figure~\ref{fig:staircase}):

\paragraph{Measured cutoffs span 22 months} (2023-12 to 2025-10) across the
twelve recall-capable rows, under one protocol and one panel. Any multi-model
comparison that fixes a single calendar placebo window is therefore
comparing different memory regimes across models.

\paragraph{Against vendor-reported dates the measurement runs early.} Where
an official date exists (7 of 12 recall-capable rows), the deviation is
one-sided: \textbf{five early}, one exact, one late. Early by six months
(GPT-5.5: 2025-06 vs.\ 2025-12), seven to eight (both Gemini 2.5 tiers
vs.\ 2025-01), three (Claude Sonnet 5 vs.\ its ``reliable'' tier 2026-01),
and four (Claude Haiku 4.5 vs.\ 2025-02); exact on Llama 3.1; and
\emph{late} by one month on GPT-5.4-mini (measured 2025-09 vs.\ 2025-08),
which we do not read as a reverse bias --- one month is inside the
resolution of our own threshold sweep (\S\ref{sec:limitations}), and that
model's collapse ramp straddles the official date. Anthropic's two-tier
disclosure (reliable/training) itself documents that the behavioral
boundary is not the data boundary; our measurements fall earlier than even
the conservative tier. Where no official date exists (DeepSeek, Qwen,
Moonshot) the only available comparison is to third-party trackers, which
conflict with each other and with the measurement in both directions ---
DeepSeek's measured 2025-07 \emph{postdates} circulating estimates. That
two-sided spread is a property of trackers, not of vendor disclosure, and
we keep the two comparisons separate. The behavioral locator, not
disclosure or tracker consensus, is the auditable quantity. This extends the effective-vs-reported gap of
\citet{dateddata2024} and the declared-vs-detected divergence of
\citet{llmlagbench2025} from the freshness level to the in-task level.

\paragraph{Post-cutoff windows must be model-relative.} For five rows the
measured cutoff falls \emph{inside} the calendar post-cutoff window that a
designer would naively draw, so the placebo is recomputed on each model's
own post-boundary months, with both computations frozen and released per
model (Table~\ref{tab:rows} prints the model-relative
placebo\textsuperscript{m}; its REC and LAP post columns print
calendar-window values). For the placebo the five corrections happen to be
small; for the recovery probe the same correction flips the reading
outright --- one model ``recovers'' the dates of 69\% of its convergent
calendar post-cutoff probes (11/16 over the 17-month window), but 9 of
those 17 months are still inside its measured memory; on its
model-relative window, where recall is provably zero, recovery drops to
38\% (3/8), which is the number that can be interpreted as pure inference. A
column label like \emph{post-cutoff} is only as meaningful as the boundary
it is relative to. The same logic applies to any train/test split or ``post-cutoff
safe'' deployment claim: safety is relative to the measured, task-level
boundary, and \S\ref{sec:patterns}'s memory-without-trigger rows show the
converse --- a date inside the memory window is not automatically
dangerous either. Freshness-level locators
\citep{llmlagbench2025,cutoffbench2025} and this in-task locator are
complements: the former audit the model in general, the latter audits the
evaluation you are actually running.

%% file: sections/07_invariances.tex
\section{Protocol Invariances and Operational Requirements}\label{sec:invariances}

A benchmark of behavioral reflexes is only portable if its numbers survive
the details of serving. We tested this directly where it is cheapest to get
wrong, and codify the results as requirements.

\subsection{Quantization is not neutral (BM-2a)}\label{sec:quant}

We re-ran the full arm matrix on the same weights (Qwen3.6-27B) under three
serving precisions, with a preregistered mutual-CI stability criterion
against the frozen FP8 reference row (E2 $+19.3\pp$ [10.8, 28.5]):

\begin{itemize}
\item \textbf{AWQ-INT4: stable.} E2 $+17.2\pp$ [6.6, 28.0]; E3 and the
measured cutoff also stable (within one month).
\item \textbf{BF16: unstable.} Pooled over two full reps, E2 falls to
$+5.6\pp$ $[-2.3, +13.4]$ --- the CI no longer contains the reference
point estimate, failing mutual containment on double data. E3
($+15.4\pp$ [6.3, 24.3]) and the cutoff (identical month) remain stable.
\end{itemize}

Decomposition localizes the entire divergence to the \emph{masked} arm:
the D-arm crisis--calm gap is tier-invariant, while the BF16 M-arm gap is
positive ($+6.6\pp$ [0.1, 12.9], pooled) where the FP8 reference is null
($-3.5\pp$). Under the preregistered decision rule for the follow-up rep,
this passes the letter of the ``masked arm leaks at full precision''
criterion --- but only at the evidence boundary (hairline CI, strong
between-rep heterogeneity), and we report it as boundary evidence, not a
mechanism claim. The protocol-level conclusion does not depend on the
mechanism: \textbf{audit rows must pin and report the serving quantization,
and comparisons across quantizations of ``the same model'' are not valid
without a stability check}. Existing multi-model sweeps that mix official
APIs with community-quantized checkpoints inherit this silently. This is
the audit-metric face of a broader finding in the quantization-evaluation
literature: compression can preserve aggregate accuracy while flipping
individual predictions \citep{dutta2024accuracy}, with effects uneven
across tasks and behavioral dimensions \citep{li2024evaluating,
fu2025quantized}; AWQ-INT4's stability here is consistent with its
weight-only, activation-aware design \citep{lin2024awq}.

\subsection{The reasoning regime can disable a probe}\label{sec:thinking}

On Kimi-K2.6 (provider-locked thinking, temperature locked to 1), the
date-recovery probe fails to converge: 256 of 258 dates exhaust a
16,384-token budget reasoning without emitting an answer, reproduced across
JSON-mode on/off and batch/direct serving; the two convergent dates used
$\sim$7k tokens. The arms and the LAP probe converge at full strength ---
so date \emph{assertion} and date \emph{inference} dissociate even at the
compute level, and the REC column is reported as unmeasurable rather than
extrapolated from $n=2$. Separately, during probe calibration we observed
that compressing a model's thinking budget can flip its recall-answer
distribution outright; cross-model comparisons must therefore pin the
thinking regime per model and disclose provider locks. Batch serving also
truncated reasoning-heavy generations at a far higher rate than direct
serving on the same requests --- batch output budgets need independent
headroom.

\subsection{Parsing, sampling, and compliance policy}

\textbf{Parser policy is part of the measurement.} Our v2 parser adds one
uniform recovery (retry after trailing-comma normalization). It changes
nothing on 13 of 15 rows but recovers 100 generations on one model
(shifting its E2 by $+5.8\pp$) and 10 on a second ($-0.2\pp$) --- a pure
artifact of strict-JSON scoring, not of the models' hindsight. Report the
parser version; apply changes
uniformly; disclose per-row deltas. We failed the ``uniformly'' half of our
own rule on the \emph{probe} side and disclose it rather than restate the
cells: two probe-answer parsers coexist in our runners, and the stricter
one --- first whitespace token, which voids schema-adjacent replies like
\texttt{\{"answer": "unknown"\}} --- was applied to the two Gemini rows
only. It voids \textbf{25.7\%} of one row's probe answers and 3.4\% of the
other's against \textbf{0.0\%} everywhere else, and voided answers sit in
LAP's denominator, biasing those two rows' LAP down. Re-parsing them under
the permissive rule changes none of the 8{,}818 answers the strict rule
already accepted (3{,}834 on one row, 4{,}984 on the other), and moves
LAP\textsubscript{pre} .337~$\to$~.430 and
.922~$\to$~.954 and $\delta$ from $-0.266$ to $-0.250$ ($t=-2.09$), leaving
$\delta$'s sign and significance and both measured cutoffs unchanged at the
reported threshold. The lesson is the rule itself: a probe parser is a
measurement instrument, and mixing two of them across a panel makes the
LAP column non-comparable between rows even when every row is individually
correct. \textbf{Sampling locks}: providers that
reject non-default sampling get preregistered exceptions recorded in the
row (one row runs at provider defaults). \textbf{Retrieval off.} Several 2026-generation endpoints enable
retrieval or tool use by default; every metric here presupposes pure
parametric behavior, and the post-cutoff placebo's premise (no memory to
trigger) is void under live retrieval. Rows must pin retrieval/tools off
and record the setting; all fifteen rows here are plain completions.
\textbf{Compliance and identifiability gates}: rows
with low schema compliance keep their VALID number and lose only the
metrics that stop being identified (\S\ref{sec:metrics}) --- and the three
small models gated by the recall-hit-rate condition are the existence proof
that ungated columns would report artifacts (never-collapsing
``cutoffs'').

\subsection{Cross-domain check (GD-2)}\label{sec:crossdomain}

A reduced 65-date audit on 10-year Treasury direction, three models, one
rep, direction semantics pinned before running: Claude Haiku 4.5 E2
$+63.2\pp$ [45.9, 80.0] and E3 $+62.9\pp$ [24.6, 98.1]; DeepSeek v4-flash
$+42.3\pp$ [11.7, 71.7] and $+64.3\pp$ [23.6, 104.2]; GPT-5.4-mini E2
$+20.6\pp$ [7.3, 34.1] with E3 positive but not significant
($+13.8\pp$ $[-5.3, +28.2]$) under the reduced window's structurally wide
transplant CI (the preregistration's prediction failed partially on this
model; reported as-is). Bond-domain effects exceed the same models' equity
effects on five of the six arm-level contrasts (the exception is
GPT-5.4-mini's transplant, itself not significant under the reduced
window), consistent with the flight-to-safety asymmetry in the companion
paper. Total marginal cost of the cross-domain check: \$4.89.

%% file: sections/08_maintenance.tex
\section{Maintenance and Versioning}\label{sec:maintenance}

\paragraph{Frozen collection contract, versioned scoring, rolling window.}
The data-collection contract (arms, probes, primary estimands, bootstrap
conventions) is frozen by hash and does not change with new entrants; the
scoring specification is versioned---a correction to an identifiability gate
or parser creates a new scoring version and regenerates every affected row
(below). The post-cutoff placebo window rolls forward
for \emph{new} audits as calendar time passes: new months are appended to
the extension panel, and a new row's model-relative window is derived from
its measured cutoff at audit time. Published rows keep the windows they
were frozen with; re-auditing an existing model under an extended window
produces a new row version (below). The benchmark therefore does not
silently expire as models' knowledge advances.

\paragraph{Row versioning.} Each row records its protocol tier (full or
reduced), rep count, sampling configuration, serving quantization,
thinking-regime setting (including provider locks),
retrieval/tool setting, parser version, and adapter; re-audits of the same model under a new serving
configuration are new rows, not overwrites (the quantization result of
\S\ref{sec:quant} is why). Vendor-reported cutoffs are stored alongside
measured ones with the date the vendor documentation was read, since that
documentation changes: \texttt{outputs/bench/vendor\_cutoffs.json} pairs
each row's measured cutoff with the vendor value, the disclosure type
(official, official two-tier, or none), and the check date, and records
the signed month difference. Models whose vendors publish nothing are
marked as such rather than filled in from third-party trackers.

\paragraph{Adding a model.} A new entrant requires an API key, one smoke
run with manual inspection (2 dates $\times$ 4 arms, gated in the
preregistration), the full run, and one analysis command that emits the row
JSON. The leaderboard table, CSV, and both papers' table files regenerate
from row JSONs by one script whose \texttt{--check} mode acts as a drift
gate; display metadata (names, generation assignment, disclosure marks)
lives in a single registry file. Data-dependent additions after a table
freeze are permitted but must be marked as such on the row and disclosed
(the disclosed addition Qwen3-30B-A3B, note\textsuperscript{n} in
Table~\ref{tab:rows}, is the worked example).

\paragraph{Retirement and correction.} Rows are never deleted; superseded
rows (deprecated endpoints, corrected artifacts) are flagged with a reason
and kept in the history. Corrections that change any published number
require a regeneration of all derived tables --- which the drift gate makes
mechanical --- and a dated changelog entry.

%% file: sections/09_limitations.tex
\section{Limitations}\label{sec:limitations}

\textbf{One primary carrier task.} The profile rows are measured on a
single monthly U.S.\ macro panel; the cross-domain check
(\S\ref{sec:crossdomain}) is a reduced-protocol audit on one additional
domain and three models, not a second full leaderboard. Task-conditionality
is real --- the companion paper's dissociation results differ from
single-stock evidence --- so rows should be read as ``on this task
class.'' \textbf{Masked-arm estimates are lower bounds}: masking removes
the trigger, not the knowledge. \textbf{Black-box probes approximate.}
Without logprobs, LAP is a frequency estimate from 20 samples (10 on the
reduced tier and on the disclosed post-freeze row); the collapse-threshold
convention (LAP $> 0.1$) is frozen but not canonical. Post-freeze
sensitivity checks bound both choices: re-sweeping the threshold across
$[0.05, 0.2]$ moves the measured cutoff by at most one month on 9 of the 12
rows with a defined cutoff (at most nine months anywhere, on Kimi-K2.6), and
on three re-probed rows doubling to 40 samples moves no cutoff by more
than one month --- while re-probing the same rows at temperature 0.3/0.7
moves one cutoff by four months, so the sampling regime, not estimator
resolution, is the binding convention (\S\ref{sec:invariances}).
\textbf{Coverage is uneven by construction.} REC is
unmeasured on two rows (one non-convergent reasoning regime; one
reduced-tier row whose probe yields almost no convergent answers, below the
$n \geq 10$ gate); one
row's placebo arms were not generated (cost decision); the cost ledger has
disclosed gaps. Reduced-tier rows carry systematically wider CIs by
construction (65 dates, fewer reps); comparisons that lean on them are
correspondingly weaker.
We prefer explicit holes to imputed values. \textbf{Breadth is not causal
depth.} The leaderboard identifies patterns (generation-not-scale,
both-sign $\delta$); the causal anatomy behind them rests on the companion
paper's three-model core, and the within-vendor Qwen contrast partially
controls for vendor and active-parameter scale within one MoE family, but
does not hold total architecture or corpus period fixed (\S\ref{sec:patterns}). \textbf{Rows profile the generation layer.} An audit row measures
generation-level behavior; how much profiled hindsight survives into
decisions depends on the surrounding pipeline and aggregation architecture
(the companion paper measures that propagation directly), so deployment
relevance is pipeline-conditional. \textbf{Direction-only
readout.} The behavioral field is directional (bearish share); magnitude
and calibration hindsight are out of scope. \textbf{English-only prompts}
on a U.S.-centric panel; corpus-culture effects on the trigger are an open
question the leaderboard's vendor spread hints at but cannot settle.

%% file: sections/10_ethics.tex
\section*{Ethics and Dual Use}

HindsightBench localizes where a model retains behaviorally active
outcome knowledge. The same rows that warn a deployer could, in
principle, guide an adversarial user toward the model whose memory of a
target period is strongest --- for instance, to make a backtest look
better than it is. We judge the marginal risk limited: an audit row
reveals dated, direction-level, aggregate knowledge already implied by
training-data coverage, not private information, and the protocol's
primary deployment effect is to invalidate misleading backtests rather
than to enable them --- the same measurement that finds residual
hindsight is the disclosure that discounts it. The released protocol
pins retrieval off, freezes prompts and configurations before
measurement, and ships transcripts for independent review; the panel is
public-domain government macro data (ALFRED vintages) with no personal
information. Cost disclosure (\S\ref{sec:cost}) keeps the audit
reproducible by independent parties rather than only by the vendor being
audited.